\documentclass[conference]{IEEEtran}
\IEEEoverridecommandlockouts
\usepackage{cite}
\usepackage{amsmath,amssymb,amsfonts}
\usepackage{algorithm}
\usepackage{algorithmic}
\usepackage{graphicx}
\usepackage{textcomp}
\usepackage{xcolor}
\def\BibTeX{{\rm B\kern-.05em{\sc i\kern-.025em b}\kern-.08em
    T\kern-.1667em\lower.7ex\hbox{E}\kern-.125emX}}
\begin{document}

\title{Exploring Biases and Prejudice of Facial Synthesis via Semantic Latent Space
}

\author{
    \IEEEauthorblockN{Xuyang Shen \ \ Jo Plested \ \ Sabrina Caldwell \ \ Tom Gedeon} 
    
    \IEEEauthorblockA{
        \textit{Research School of Computer Science}, \textit{the Australian National University}\\
        Canberra, Australia \\
        first.second@anu.edu.au
    }
}

\maketitle

\begin{abstract}
Deep learning (DL) models are widely used to provide a more convenient and smarter life. However, biased algorithms will negatively influence us. For instance, groups targeted by biased algorithms will feel unfairly treated and even fearful of negative consequences of these biases. This work targets biased generative models' behaviors, identifying the cause of the biases and eliminating them. We can (as expected) conclude that biased data causes biased predictions of face frontalization models. Varying the proportions of male and female faces in the training data can have a substantial effect on behavior on the test data: we found that the seemingly obvious choice of 50:50 proportions was not the best for this dataset to reduce biased behavior on female faces, which was 71\% unbiased as compared to our top unbiased rate of 84\%. Failure in generation and generating incorrect gender faces are two behaviors of these models. In addition, only some layers in face frontalization models are vulnerable to biased datasets. Optimizing the skip-connections of the generator in face frontalization models can make models less biased. We conclude that it is likely to be impossible to eliminate all training bias without an unlimited size dataset, and our experiments show that the bias can be reduced and quantified. We believe the next best to a perfect unbiased predictor is one that has minimized the remaining known bias.

\end{abstract}


\section{Introduction}

During the last decade, deep learning has demonstrated its strong ability in various classification and regression problems. Deep learning algorithms are not only favored by many researchers but also popular in industrial applications. Nowadays, deep learning applications are common in real life, such as language translation in Google and face recognition in security. An effective application will make our life more convenient; on the other hand, a problematic application involving biased decisions can greatly reduce our experience. In deep learning, biases and prejudice are defined as differential decisions being made toward different groups and individuals based on their features or characteristics~\cite{b1}. 

Biases in datasets and pre-trained models can lead to biased stereotypes in general public~\cite{b2}. For example, before gender-specific translation, the Google translate system only matched masculine forms to the term 'doctor' and feminine forms to the term 'nurse' when translating English to Turkish~\cite{b2}. There also exist similar issues in facial synthesis. In the middle of 2020, a Twitter hotspot arose discussing the skin-color biases of the PULSE photo up-sampling model, where it recovers any human faces with different skin colors into white skin.  

Bias studies in different areas of deep learning are unbalanced. Several researchers emphasized that the area of bias in classification algorithms is well researched, but little attention has been paid to the study of bias in generative models~\cite{b1,b7}. Paying attention to learning algorithms' bias should also be an obligation for computer science researchers from both ethical and technical perspectives. Based on these motivations, we extend our previous work on facial frontalization by exploring the biases from a semantic latent space perspective~\cite{b3}. Facial frontalization refers to recovering the frontal face from a side-pose image, and belongs to the novel view synthesis domain. Two base frameworks in facial frontalization from previous research are Pix2Pix (Pixel to Pixel generative adversarial networks) and Cycle-GAN (Cycle-Consistent generative adversarial networks)~\cite{b4, b5}. The current state-of-the-art models are Pairwise-GAN and FFWM (Flow-based Feature Warping Models), and use different databases~\cite{b3, b6}. We selected Pix2Pix and Pairwise-GAN for our analysis. 

The purpose of our work is to fill in a gap in the field of group biases study associated with facial databases used to train GAN models. From this baseline, we conducted our experiments in several stages: First of all, we observed the biased behaviors in facial frontalization models from the gender biases perspective. Then, we investigated the causes of the biases following guidelines~\cite{b1, b7, b8}. Further, we plan to extend the findings from facial frontalization models to general face generation conditional-GANs.

Our main findings in this project are:
\begin{itemize}
\item There are two kinds of biased behaviors for representative models in facial frontalization, including failure in generations and generating incorrect gender. The former refers to the situation where the generated frontal face loses partial or all facial features; the second means the gender of generated frontal faces does not match the gender of the side-pose source faces.
\item Data biases lead to biased behaviors of GANs, which means that the models are vulnerable to biased datasets, and supports our initial hypothesis. In addition, only some layers in Pairwise-GAN are particularly vulnerable to biased datasets.
\item Modifying the generator architecture of Pairwise-GAN by cutting off the skip connection, leads to the model behaving in a less biased fashion. As a result, one biased behavior (failure in generation) can be largely eliminated.
\end{itemize}

\section{Related Work}

\subsection{Biases and Prejudice in Deep Learning}
An increasing attention to the bias and prejudice is emerging in deep learning~\cite{b1, b7, b8, b9}. Specifically, significant research outcomes have been achieved in study of bias in language translation ~\cite{b2, b12} and face recognition~\cite{b13, b14, b15, b16}. The biases in deep learning indicate the decision made by algorithms has different tendencies in each group or individual. They easily occur at any stage of a deep learning life-cycle, including data collection, model development, and entire system deployment~\cite{b1, b7, b8}. \textit{Historical bias} is the already existing unfairness in the data even if sampling with a perfect selection. This type of prejudice resulting from either social or natural factors~\cite{b7, b8}. \textit{Representation bias} occurs from sampling data from different populations, which can lead to a severe unfairness of domain-adoption algorithms. It is challenging to achieve high diversity in a database. ImageNet as the largest current dataset still lacks geographical diversity where the majority of images are from North America and Western Europe~\cite{b1,b17}. \textit{Measurement bias} describes the additional prejudice in the collected data when choosing the feature or labels through subjective ideas~\cite{b7}. \textit{Historical bias} occurs when the same model is trained, deployed, or pre-trained across different groups of samples. The biases from each group will be aggregated~\cite{b7}. \textit{Evaluation bias} is caused by unfair evaluation data and biased evaluation algorithms~\cite{b7}. As a result, a biased model will be published. \textit{Deployment bias} refers to the situation where the model is applied into different scenarios from its original design and intention~\cite{b7}. 

In terms of domain perspectives, one area of machine learning has received particular attention from the research community~\cite{b1}. The percentage of bias study in classification is higher than any other domain. In representation learning, fairness research is still limited, especially in group fairness. Moreover, another challenge in fairness research is equality and equity~\cite{b1}. Addressing equality was well developed, in which it requires each group to receive the same amount of resources and attention. On the other hand, little research has been achieved in equity, which requires each group to succeed by given appropriate resources. For instance, researchers were conscious of gender classification biases but still paid little attention to eliminating unfairness. This is one of our key motivations in this research, to fill this equity gap.

\subsection{Face Frontalization}

Face frontalization synthesizes the front face from input of either a left or right side-pose face. In practice, several works showed that frontal synthesis significantly improves the stability of face recognition on side-pose images~\cite{b10, b11}. At early stages in face frontalization, researchers tried to use a variational auto-encoder (VAE) to generate frontal face and improve its performance in this task~\cite{b18, b19, b20, b21}. With the publication of TP-GAN (Two Pathways Generative Adversarial Networks) and CR-GAN (Complete Representations Generative Adversarial Networks), the research on face frontalization shifted into conditional-GANs~\cite{b10, b22}. Since the high achievement in image translation by Pix2Pix (Pixel to Pixel Generative Adversarial Networks) and CycleGAN (Cycle-Consistent Adversarial Networks)~\cite{b4, b5}, recent publications in face frontalization were based on these two frameworks~\cite{b3, b6, b11, b28}. The current state-of-the-art model in face frontaliztaion using the Color FRET database is Pairwise-GAN (Pairwise Generative Adversarial Networks), constructed by pair generators and a PatchGAN as the discriminator~\cite{b3}. Compared to other models, the authors proposed to split two domains (left pose and right pose) to synthesize faces. The generator of Pairwise-GAN was also developed from Pix2Pix by utilizing two U-Nets. $G_{left}$ is only responsible for the left domain images, while $G_{right}$ takes charge for the other domain.

\subsection{Semantic Latent Space}

The latent space in GAN is the random noise sampled by the generator, which is popular in exploring semantic latent space~\cite{b23, b24, b25}. In addition, several papers were published on latent space in style transformation based on CycleGAN architecture~\cite{b5, b26, b27}. With the in-depth analysis of latent space, one major branch of domain mappings was developed into a combination of content transfer and style transfer~\cite{b26}. Compared to the one-step transfer in CycleGAN, there are two steps in the novel solution. Mapping two different domains into a shared semantic latent space are the prior stage, followed by the style transfer from intermediate content space into the target domain. A similar idea was also proposed by~\cite{b27} recently, which discovered that the semantic latent space has time-varying factors and permanent factors during the style transformation on time-specific problems. In~\cite{b25}, the authors present work on GAN dissection from intervention and dissection stages, which can independently control the object in generated images through intervention in the latent space.

\section{Methodology}

\subsection{Dataset}

We chose color FERET as our primary dataset, involving 11,338 facial images from 994 individuals. There are at least $5$ different angles for each individual recorded in one time session provided in the database. Before the experiments, we analyzed the historical bias and representation bias~\cite{b7}. The historical bias in this database cannot be accurately identified since the background of FERET as described on its official website is limited. However, some existing social issues in the dataset are inevitable as addressing such issues are not mentioned as being considered in its construction. In terms of representation bias, the object group in the Color FERET database is "ordinary people". Among 994 individuals, there are 594 males and 402 females, thus the ratio between males and females is around $6 : 4$. Although all images were captured in the United States, the Color FERET database still has a high diversity regarding human skin colors. With these properties, the representation bias is significantly reduced. 

In~\cite{b3}, we published the first version of pre-processing FERET data through MTCNN, which they did not achieve ideal states as MTCNN frequently fails in detecting side-pose faces. We selected Face++ (a commercial face analysis API) to crop and resize images. Furthermore, only left and right images around $67.5$ degrees were filtered out so as to minimize the noise caused by face angles. After the data pre-processing, there are 4,982 images in total, being 2,491 pairs (one front, one side pose). To reduce the distribution collision between training-set and test-set, we selected one time session among $14$ session to become the test-set, where the chosen session has the largest style difference compared to the others. Therefore, the test-set has 314 images (157 pairs), accounting for $6.3 \%$ of the total data samples. 

Gender bias exploration is the main target for this project. Six training-sets with different percentages of male and female images were constructed. The exact numbers of images are shown in Table~\ref{dataset}. Note that the female images in the dataset are not abundant, so we reduced the amounts of male images into $47$.

\begin{table}
\centering
\caption{Detail information of training-sets and test-set}
\label{dataset}
\begin{tabular}{c|ccc}
\hline\hline
\textbf{Name}          & \textbf{Male Amounts} & \textbf{Female Amounts} & \textbf{Ratio} \\ \hline
Train set 100  & 1454                           & 0                                & 10 : 0                          \\
Train set 91   & 1454                           & 162                              & 9 : 1                          \\
Train set 82  & 1454                           & 364                              & 8 : 2                          \\
Train set 73  & 1454                           & 624                              & 7 : 3                          \\
Train set 64   & 1320                           & 880                              & 6 : 4                          \\
Train set 55   & 880                            & 880                              & 5 : 5                          \\ \hline
Test set & 47                            & 47                               & 1 : 1                               \\ \hline\hline
\end{tabular}
\end{table}

\subsection{Explore Gender Biases in Face Frontalization}
\label{sec:exp1}

Since little research has been completed in generation models from the group perspective ~\cite{b1} and the gender bias is an important characteristic of human groups, we began our experiment by analyzing the gender bias in face frontalization models.

\begin{figure}[h]
\centerline{\includegraphics[width=0.9\columnwidth]{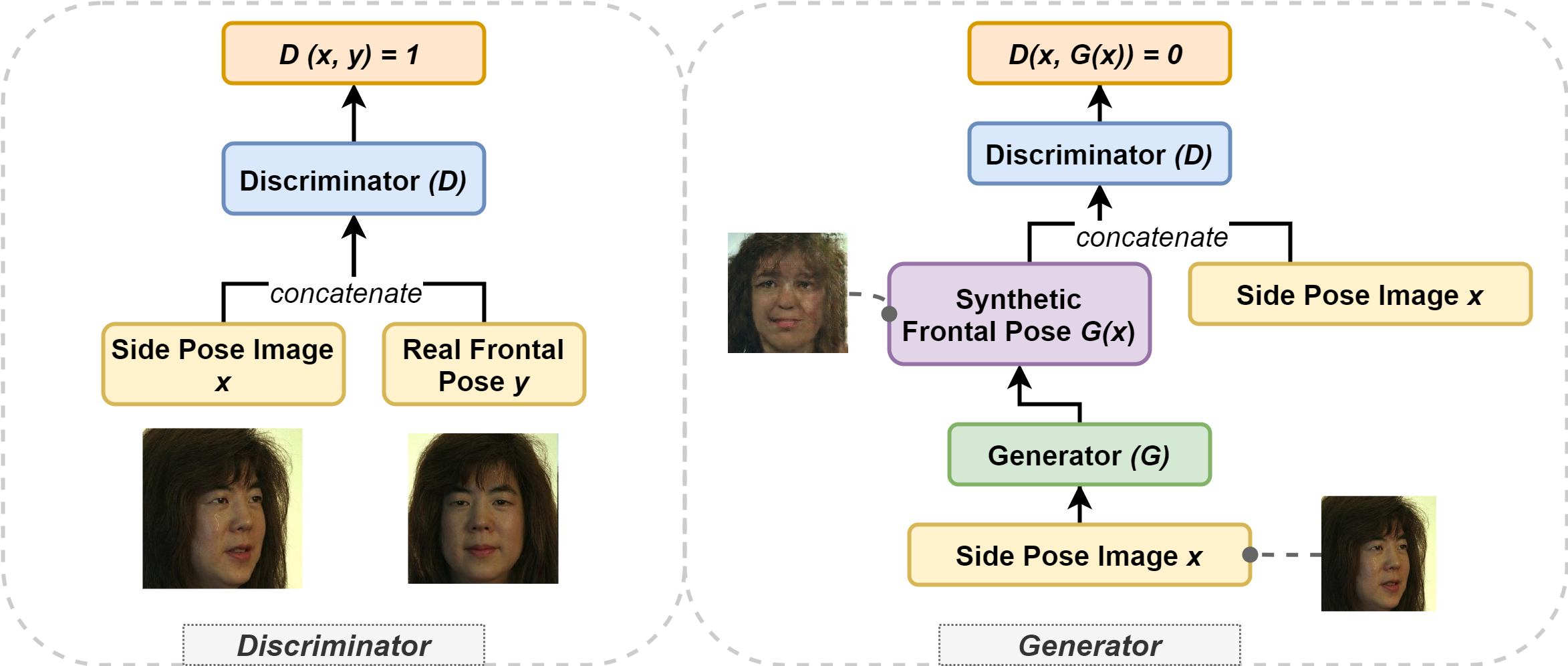}}
\caption{Pix2Pix Framework}
\label{pix2pixflow}
\end{figure}

\begin{figure}[h]
\centerline{\includegraphics[width=\columnwidth]{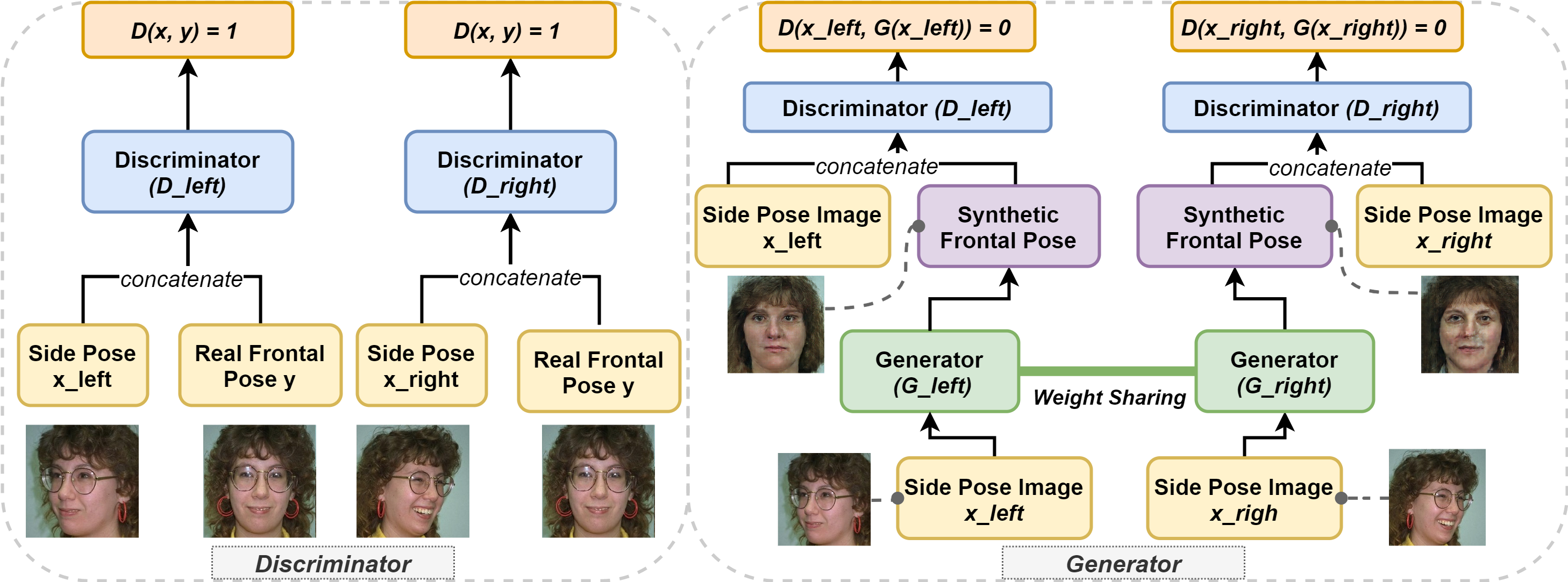}}
\caption{Pairwise-GAN Framework}
\label{pairwiseflow}
\end{figure}

\textbf{Model Architecture: } Pix2Pix and Pairwise-GAN were selected for our experiments, where the first is the baseline model in facial synthesis and the latter one is the current state-of-art model on the Color FERET database~\cite{b3,b4}. The visualizations of forward passing in Pix2Pix and Pairwise-GAN are shown in Fig~\ref{pix2pixflow} and Fig~\ref{pairwiseflow}. The generator $(G)$ in Pix2Pix is designed to map both left and right domains $(X)$ into the frontal domain $(Y)$, while there are two generators $(G_{left}, G_{right})$ in Pairwise-GAN responsible for the left or the right domains $(X_{left}, X_{right})$, respectively. Both of these models utilize Patch-based discriminators $(D)$~\cite{b4}. The loss functions of Pix2Pix $ (\mathcal{L}_{pix2pix})$ were set as adversarial loss and mean absolute loss, with suggested ratio $\lambda_{Adv} : \lambda_{L1} = 2 : 0.5$. Regarding Pairwise-GAN $(\mathcal{L}_{Pairwise-GAN})$, we follow the experimental results in~\cite{b3} to use adversarial loss ${\lambda_{Adv}}'$, mean absolute loss ${\lambda_{L1}}'$, identity loss $\lambda_{Id}$, and pair loss $\lambda_{pair}$, with a ratio of $10 : 3 : decay : decay$, where the decay weight of identity and pair Loss were $[10, 5]$ and $[10, 2]$.

\begin{equation}
    \begin{aligned}
        \mathcal{L}_{GAN} (G, D, X, Y) = & \mathbb{E}_{x,y} [log D(x, y)]   \\                                        +& \mathbb{E}_{x} [1 - log D(x, G(x))] 
    \end{aligned}
\end{equation}

\begin{equation}
    \begin{aligned}
        \mathcal{L}_{pix2pix} (G,D) = & \lambda_{Adv} \mathcal{L}_{GAN} (G, D, X, Y) \\                           +& \lambda_{L1} \mathbb{E}_{x,y} [{\| y - G(x) \|}_{1}]
    \end{aligned}
    \label{pix2pix_formula}
\end{equation}

\begin{equation}
    \begin{aligned}
        {\mathcal{L}_{GAN}}' (G, D, X, Y) = & {\lambda_{Adv}}' \mathcal{L}_{GAN} (G, D, X, Y) \\ +& {\lambda_{L1}}'\mathbb{E}_{x,y} [{\| y - G(x) \|}_{1}] \\
        +& \lambda_{Id} \mathbb{E}_{y} [{\| y - G(y) \|}_{1}]
    \end{aligned}
\end{equation}

\begin{equation}
    \begin{aligned}
        & \mathcal{L}_{Pairwise-GAN} (G_{left}, G_{right}, D_{left}, D_{right}, X_{left}, X_{right}, Y)  \\
       \ \ &= {\mathcal{L}_{GAN}}' (G_{left}, D_{left}, X_{left}, Y) \\
       \ \ &+ {\mathcal{L}_{GAN}}' (G_{right}, D_{right}, X_{right}, Y) \\
       \ \ &+ \lambda_{pair} \mathbb{E}_{x_{left}, x_{right}}[{\| G_{left}(x_{left}) - G_{right}(x_{right}) \|}_{1}]
    \end{aligned}
    \label{pairwise_gan_formula}
\end{equation}

\textbf{Experimental Setup: } Based on the two selected models and training-sets, we conducted the first experiments. First, the Pix2Pix and Pairwise-GAN were trained using $6$ training-sets with different ratios of male and female images to get $12$ pre-trained models. Then, they were all evaluated with the same predefined test-set to evaluate the bias. Moreover, to make the final results more robust, we also tried another different weight initialization method for Pairwise-GAN, producing another $6$ pre-trained models. Other hyper-parameters of these two models were left as default. All generators and discriminators were optimized with Adam; the learning rate was $0.0002$; exponential decay was enabled and set as 0.5. The training epoch for Pix2Pix and Pairwise-GAN were adjusted to $125$ and $250$, respectively, to guarantee the same amounts of data were used to train each generator. 

\subsection{Explore Latent Space on Different Predictions}
\label{sec:exp2}

In the first experiment, the test-set for trained models is located at a similar distribution to the training-set, in which the purpose was to explore the bias from normal semantic latent space. In the second experiment, the test-set to was expanded into a more broad latent space. 

\begin{figure}[t]
\centerline{\includegraphics[width=\columnwidth]{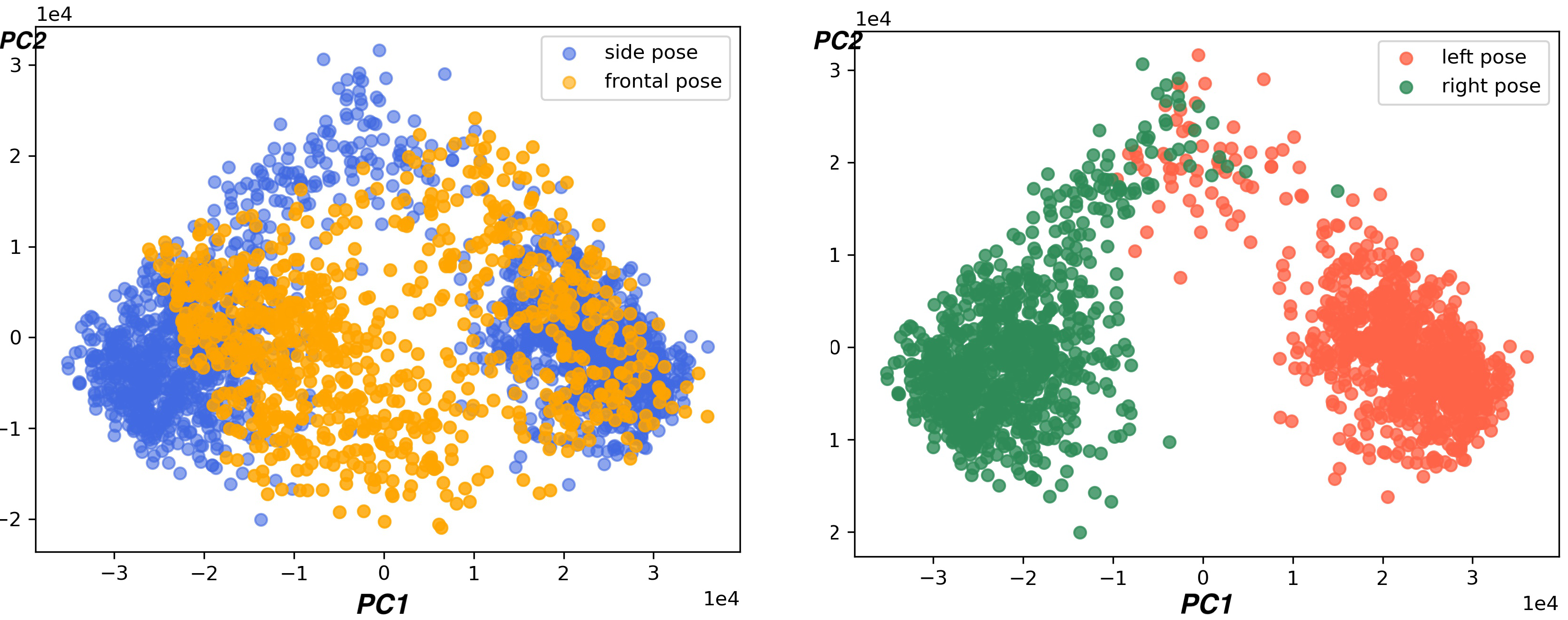}}
\caption{Left subplot: distribution of side-Pose images (blue points) and front face images (orange points). Right subplot: distribution of left-side pose (brown points) and right-side pose (light blue points). The cluster centres of left-side images and right-side images are symmetric: $(0, -25,000)$ and $(0, +25,000)$.}
\label{distribution_side_front}
\end{figure}

At the initial stage, we analyzed the mechanism of style transformation models from an intuitive viewpoint. In the left subplot of Fig.~\ref{distribution_side_front}, the distribution of each front face and side-pose face was plotted through top-2 PCA components. From the left subplot, side-pose images are aggregated on the edge of a sector, while the front faces are evenly scattered in the middle and present a diamond shape. Specifically, the distribution of two different side poses are aggregated into two clusters, where few of them are overlapped, as shown in the right subplot. Intuitively, the insightful view of domain mapping is to find the transformation function that is able to map the distribution of one domain into another one. For instance, face frontalization models are trained to map the blue points into orange points correctly. Therefore, it is meaningful to test different distribution data (latent space) for the pre-trained models to analyze the bias further. This experiment can validate whether biases can be transferred from data into trained GANs. 

\begin{figure}[h]
\centerline{\includegraphics[width=\columnwidth]{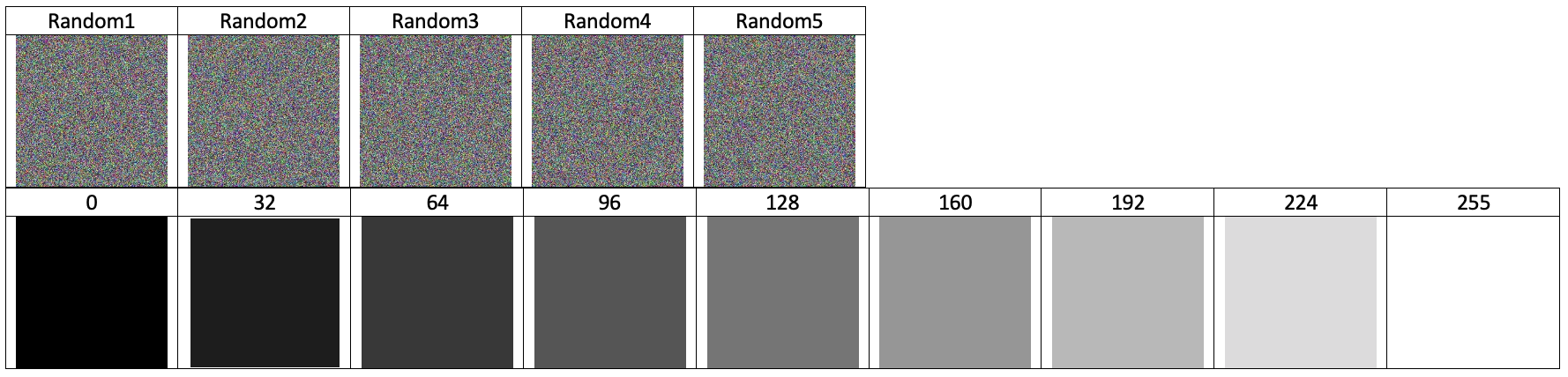}}
\caption{The pixel values in the upper row are in normal distributions. The bottom row displays nine gray-scale images where the step of nearby value is 32.}
\label{evaluationset}
\end{figure}

\begin{figure}[t]
\centerline{\includegraphics[height=3.5cm]{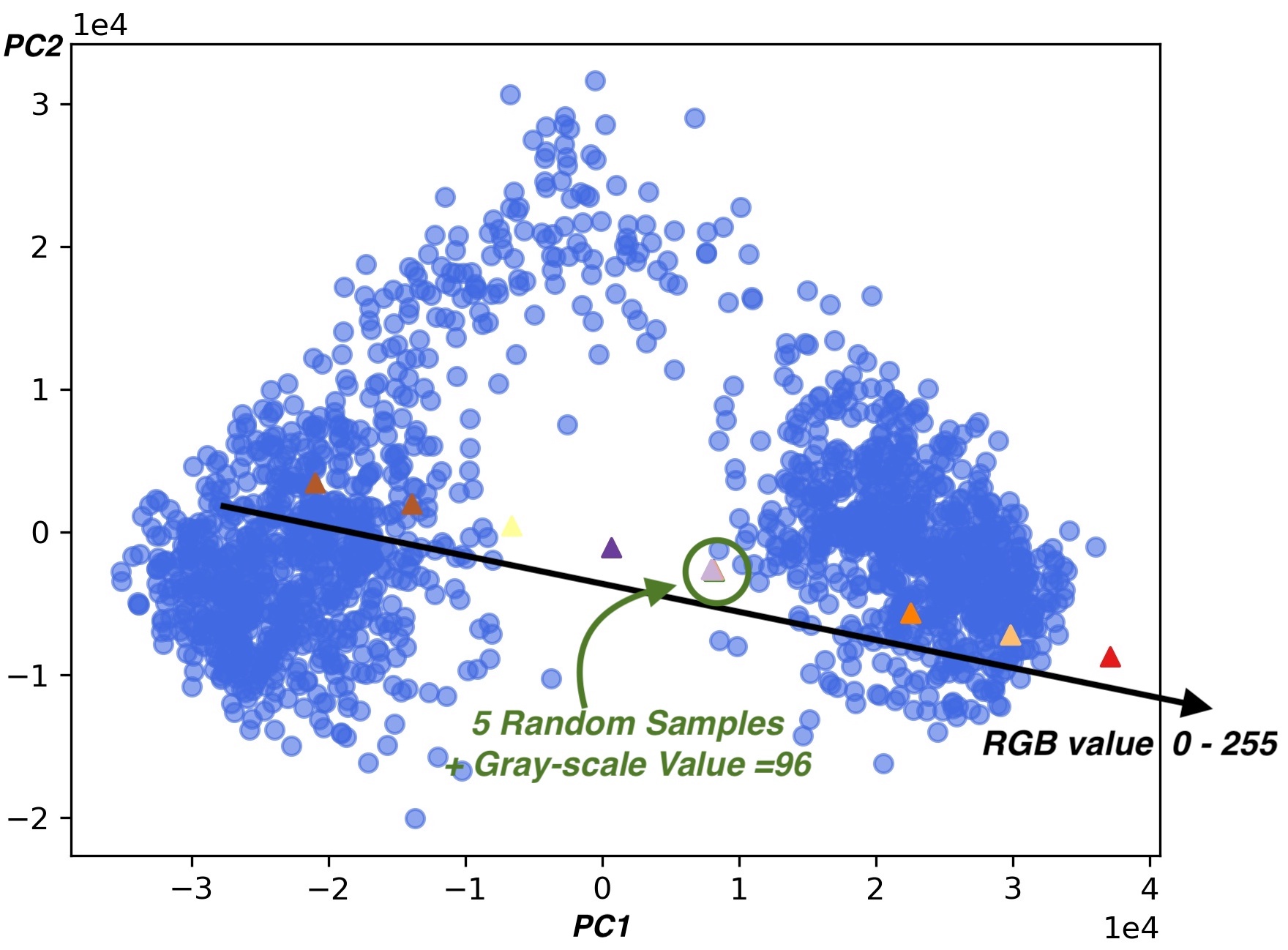}}
\caption{Through PCA transformation, the positions of nine gray-scale images are located close to a linear line. The coordinates of gray-scale value from $0$ to $255$ are: $(37145, -8729)$, $(29856, -7202)$, $(22566, -5674)$, $(7988, -2619)$, $(698, -1091)$, $(-6590, 435)$, $(-13880, 1963)$, $(-20942, 3442)$. The positions of five random-value images are overlapped into a point with location around $(8199, -2741)$.}
\label{evaluationset2}
\end{figure}

\textbf{Experimental Setup: } Since we only modified the test-set for this experiment, $18$ pre-trained models from the first experiment were selected to be tested. One test-set was sampled from random integer values in a normal distribution whose range is from $0 - 255$. Another test-set was sampled from grey-scale values, because our intent was to analyze whether different biased models were sensitive to different grey values. The images of the test-sets and their distribution are visualized in Fig.~\ref{evaluationset} and Fig.~\ref{evaluationset2}. Note that each evaluation sample for each pre-trained model will be tested more than $10$ times to obtain robust results.

\subsection{Exploring the biased layer(s)}
\label{sec:exp3}

After the previous experiments on bias exploration, we found that both Pix2Pix and Pairwise-GAN were vulnerable to the dataset bias. In the third experiment, we further analyzed the bias deeply to identify which layers were sensitive to the biased data. The purpose of this experiment was to provides a guideline on improving the stability of models to resist biased datasets. 

We investigated the variance of the latent space $(L.S.)$ to inspect the biases in layers, since it can be used to evaluate the spread range of a variable. The formula to calculate the variance of $j^{th}$ layer from $i^{th}$ model is listed below. From the distribution map, as shown in Fig.~\ref{distribution_side_front}, the side-pose domain is more sparsely distributed, compared to the frontal domain. Therefore, a hypothesis was posed that the variance should be decreasing overall. Furthermore, the encoder should aggregate the diversity of vector values since it aims to converge the data. Recovering different sample features is the objective of decoders, in which the variance should be increasing. 

\begin{equation}
    \begin{aligned}
        Var_{(L_j, M_i)} =  \mathbb{E}[(L_j(L.S._{j-1}) - \mathbb{E}[L_j(L.S._{j-1})]) ^2]\\
        where \ L.S._{0} = input, \ L.S._{1} = L_1(L.S._{0})
    \end{aligned}
    \label{variacne}
\end{equation}

\begin{figure}[htbp]
\centerline{\includegraphics[width=0.9\columnwidth]{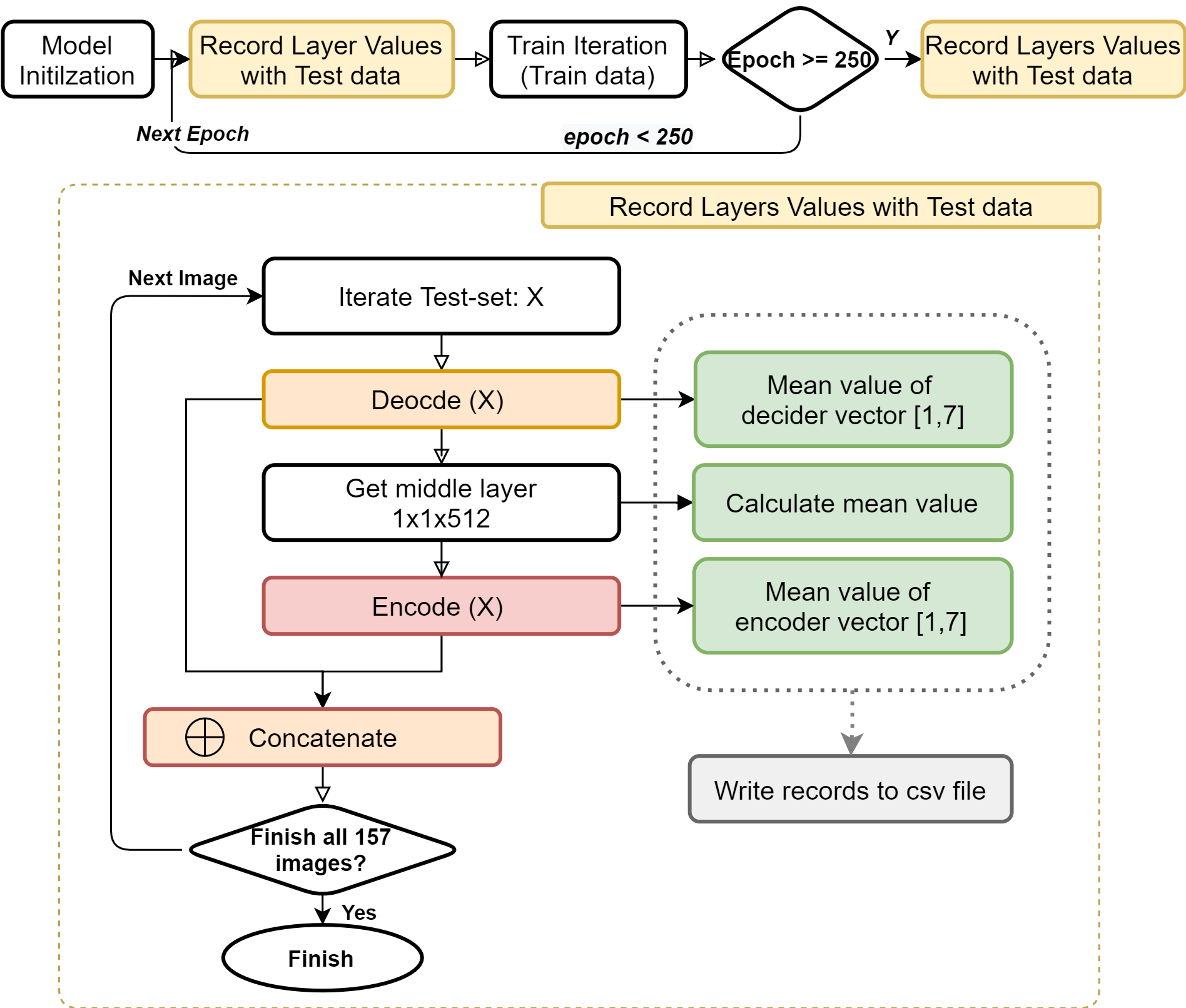}}
\caption{Top subplots describe the overall work-flow during training and prediction. The bottom subplot describes how to calculate the layer variance.}
\label{layanalysisflow}
\end{figure}

\textbf{Experimental Setup: } We selected Pairwise-GAN in this experiment as it had a better performance compared to the Pix2Pix~\cite{b3}. The hyper-parameters of Pairwise-GAN remained unchanged from the previous experiments. We enhanced the training process, adding calculation of the variance on each latent semantic vector. The entire mechanism is shown in flow chart~\ref{layanalysisflow}. We used the test-set in experiment~\ref{sec:exp1} to examine the latent vector during training and prediction. 

\section{Results and Discussion}

\subsection{Evaluation Methods}
\label{sec:evaluation}

Unfair evaluation leads to misrepresentative conclusions~\cite{b7}; thus, choosing evaluation methods is a vital step. To observe the biased behaviors of face frontalization models, we defined two measurements. The first measure is to check whether the generative model can recover a facial image with facial features (Face Recovery Rate). The second criterion is to classify whether the gender of generated images matches the gender of the ground-truth image (Gender Match Rate). In the first two experiments, there were respectively $3,768$ and $20,160$ predicted images to be examined for gender. Therefore, ~\textit{gender classification algorithms} were examined and analyzed: "\textit{PY-agender}" and "\textit{Face ++ API}". Among $1000$ front faces, the first failed to detect $31$ cases, while "\textit{Face ++}" achieved a $0$ failure rate. Furthermore, in the false positive rates, \textit{Face ++} behaved more accurately than \textit{PY-agender}. Thus, we selected \textit{Face ++} as the measurement tool. However, we note that in common gender classification algorithms, the accuracy of prediction results is unbalanced between different genders; in particular, female images can easily be classified into the male gender. This we consider as a limitation in our evaluation work, and illustrates the difficulties in achieving unbiased results.

\subsection{Gender Bias in Face Frontalization Analysis}
\label{sec:result1}

\begin{figure}[htbp]
\centerline{\includegraphics[width=0.97\columnwidth]{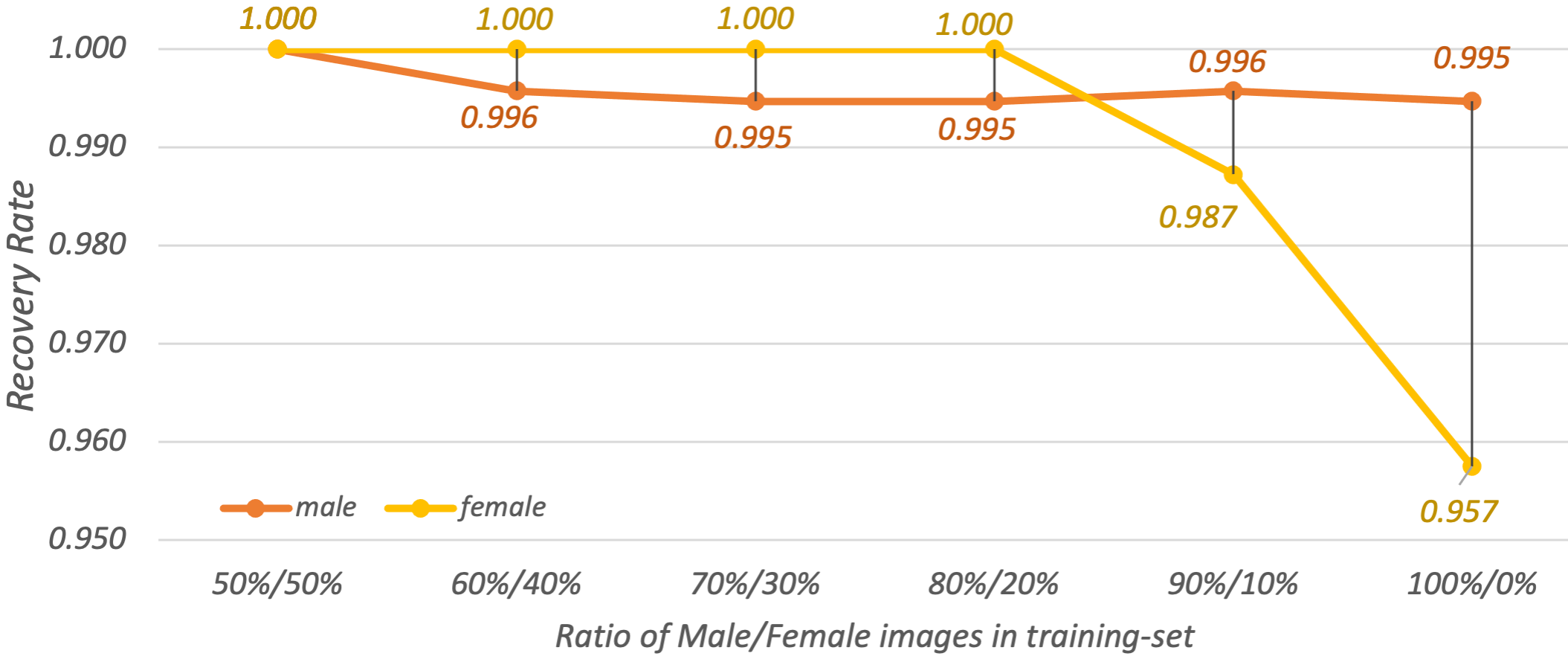}}
\caption{Face Recovery Rate of Different Pre-trained Models}
\label{generate1}
\end{figure}

\begin{figure}[htbp]
\centerline{\includegraphics[width=0.97\columnwidth]{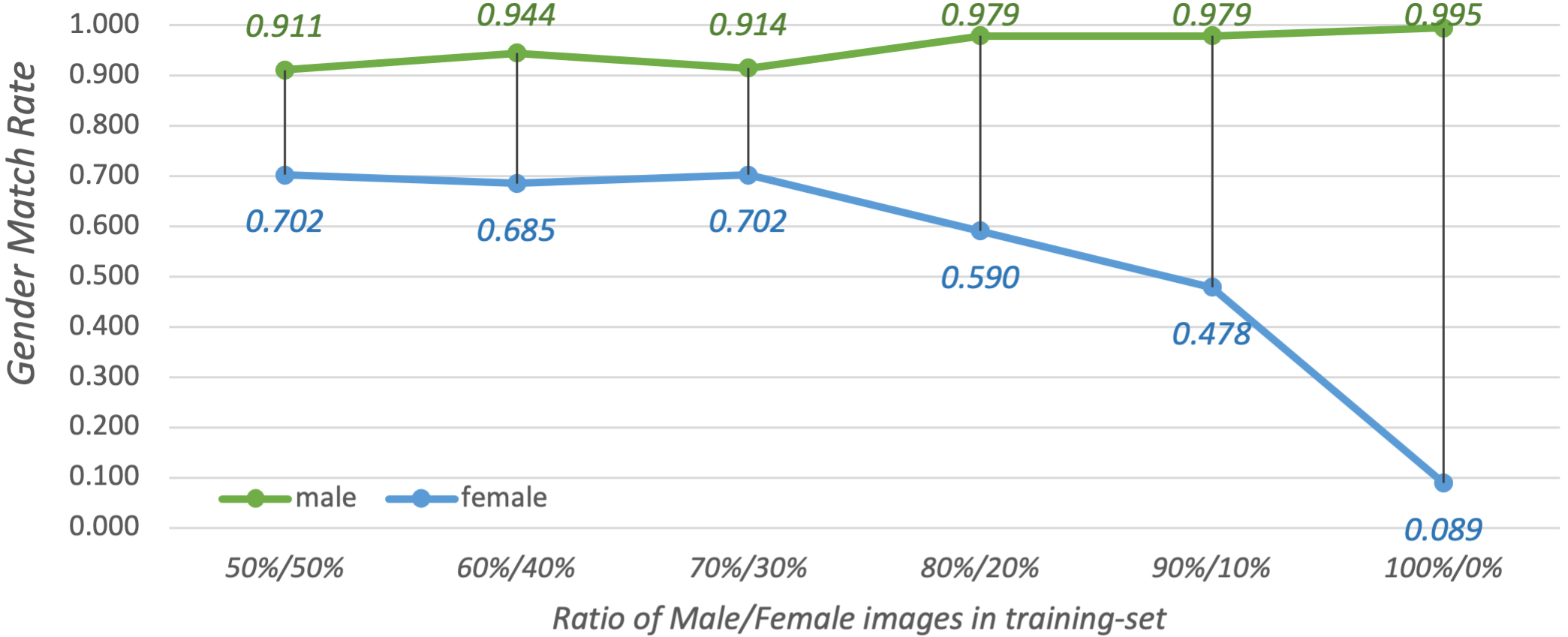}}
\caption{Gender Match Rate of Different Pre-trained Models}
\label{gender1}
\end{figure}

The first stage of the overall experiment was to explore the gender biases in two face frontalization models. In summary, Pix2Pix and Pairwise-GAN were trained on six different training-sets with different initialization strategies, producing $18$ pre-trained models. The result of our first stage experiments are shown in Fig.~\ref{generate1} and Fig.~\ref{gender1}. 

In Fig.~\ref{generate1}, the x-direction from left to right represents the descending percentage of female images in training-sets; the y-direction indicates the face recovery rate. With the ascending proportion of male images in training-sets, recovery rates of male side-pose images fluctuates around $99.6\%$. However, the recovery rates of female side-pose images trends in a decrease; primarily, when zero female images are involved in the training-set, around $4\%$ of female side-pose images cannot be used to generate a face image. From the face recovery perspective, our models show a huge difference in biased behaviors only when the percentage of female images is lower than $20\%$ in training-sets.   

The situation changes when analyzing the gender match rate. In Fig.~\ref{gender1}, the x-direction represents the descending percentage of female images in training-sets; the y-direction indicates the gender match rate. With an increasing percentage of male images in training-sets, the gender match rate of a generated frontal image from a male side-pose image fluctuates around $94\%$. On the other hand, the gender match rate from female side-pose images drops heavily down with the descending number of female images in training-sets. To be specific, if the model is trained on all-male images, only $8.9\%$ of generated female images are correct in gender. In other words, if the pre-trained model receives a female side-pose image, there is $91\%$ probability that the predicted frontal image is male. We also observe that the matching rate of female images is $20\%$ lower than male images even when models are trained on an equivalent proportion of males and females in a dataset. This is most likely caused by the incorrect gender-classified results from the classification algorithm. In summary, face frontalization models are vulnerable to extremely biased data. In particular, the most severe behavior of a biased model is to recover an incorrect gender face from a side-pose image.

\subsection{Explore semantic latent space on different prediction data}
\label{sec:result2}

Based on the first experiment, we explored the latent space more by including $5$ random noise from Gaussian distributions and $9$ gray-scale images from black to white in the test-set (Fig.~\ref{evaluationset}). Analysis strategies followed the previous experiment, which is to investigate two criteria. 

\begin{figure}[htbp]
\centerline{\includegraphics[width=\columnwidth]{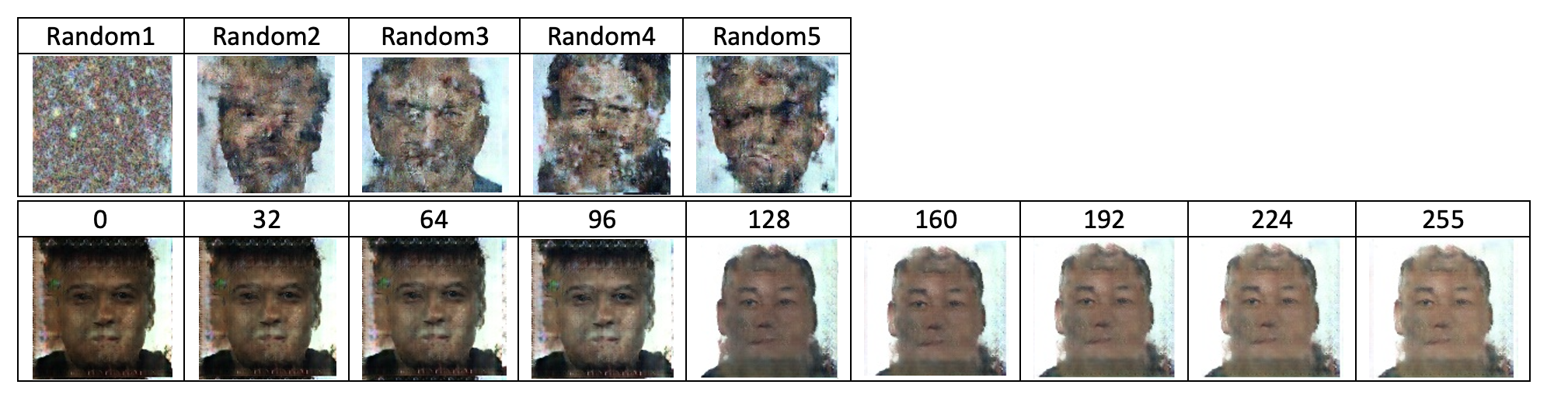}}
\caption{One Prediction Result on Test-set~\ref{evaluationset}.}
\label{emptyresults}
\end{figure}

From the face generation perspective, one series of recovered images is displayed in Fig~\ref{emptyresults}. If the input image is random noise, all no pre-trained models can generate close to human face images, and only the contour of faces can be identified. If the input images are gray-scale images, the model can generate a facial image with a high probability. Compared to light gray-scale images, it is easier for models to generate face images from dark images. In Fig.~\ref{emptyresults}, we can observe that in test images with RGB value from $0$ to $128$, the predicted images are close to one person. If the RGB value is large than $128$, the predicted results are close to another person. With ascending value in test images, generated images become lighter as well.

The second goal is to measure the gender of generated images among different pre-trained models, whose results are shown in table~\ref{trainpred} and table.~\ref{testpred}. Since the random noise cannot be recovered as human faces, they are removed from following analysis. Note that the outputs of left generator and right generator in Pairwise-GAN may exist minor difference, which leads to mixed results of gender classification. We classified these models in the last row of two tables. Table~\ref{trainpred} describes the prediction on $9$ gray-scale types from models trained on different train-sets. If the pre-trained models are trained on an equivalent number of female images and male images, a high proportion of prediction results are classified as female. With the ascending percentage of male images in training-sets, the probability of generating a male image is higher than generating a female image. This indicates the same conclusion as the first experiment, in which biases in training-sets are transferred into pre-trained models. Besides, we can also infer that unbiased train-set and extremely biased train-set result stable prediction from two generators of Pairwise-GAN. The second table.~\ref{testpred} summarizes the predicted results of pre-trained models on $9$ gray-scale test-set. For instance, when the test image is black ($RGB=0$), predicted frontal faces from $5$ pre-trained models appear both gender among $52$ test samples, where some are classified as male and others are classified as female. Images with lower RGB values (dark images) have a slightly higher probability of being generated as a female frontal face. In addition, around $90\%$ of pre-trained models recover a male image from an image with high RGB values.

\begin{table}[h]
\centering
\caption{Influence of training-set to bias of pre-trained models}
\label{trainpred}
\begin{tabular}{r|cccccc}
\hline\hline
\textbf{Male in train-set (\%)}                              & 50 & 60 & 70 & 80 & 90 & 100 \\
\textbf{Female in train-set (\%)}                            & 50 & 40 & 30 & 20 & 10 & 0   \\ \hline
\textbf{\# of pre-trained models}                            & 3  & 3  & 3  & 3  & 3  & 3   \\
\textbf{\# of test-set types}                                & 9  & 9  & 9  & 9  & 9  & 9   \\ \hline
\multicolumn{1}{c|}{\textit{Output male}}   & 5 & 14 & 15 & 15 & 27 & 27  \\
\multicolumn{1}{c|}{\textit{Output female}} & 21  & 4  & 8  & 4  & 0  & 0   \\ \hline
\multicolumn{1}{c|}{\textit{Mixed cases}} & 1  & 9  & 4  & 8  & 0  & 0   \\ \hline\hline
\end{tabular}
\end{table}

\begin{table}[h]
\centering
\caption{Influence of test-set to bias of pre-trained models}
\label{testpred}
\begin{tabular}{c|ccccccccc}
\hline\hline
\multicolumn{1}{r|}{\textbf{Test-set type}} & 0  & 32 & 64 & 96 & 128 & 160 & 192 & 224 & 255 \\ \hline
\multicolumn{1}{r|}{\textbf{\# of models}}  & 18 & 18 & 18 & 18 & 18  & 18  & 18  & 18  & 18  \\ \hline
\textit{Output male}                      & 6  & 7  & 6  & 6  & 16  & 15  & 15  & 15  & 13  \\
\textit{Output female}                    & 7  & 7  & 7  & 7  & 1   & 2   & 2   & 2   & 2   \\ \hline
\textit{Mixed cases}            & 5  & 4  & 5  & 5  & 1   & 0   & 0   & 0   & 2   \\ \hline\hline
\end{tabular}
\end{table}

The previous finding summarises that all pre-trained models are sensitive to the RGB value of input images. Therefore, we also evaluated the RGB values in training-sets. The average RGB values of entire images and facial parts are calculated, which the second measurement is able to indicate the skin color of test images. The average RGB values of male images and female images in training-sets are $117.4$ and $105.4$. Moreover, when just the faces are cropped from the images, the average RGB values of male faces and female faces in training-sets are $92.3$ and $94.5$.  This indicates that the RGB values of entire images are more important for these models than the facial part. In addition, this suggests that models in face frontalization are vulnerable to the skin and hair color of training-sets as well.

\subsection{Results on the biased layer study}
\label{sec:result3}

\begin{figure}[htbp]
\centerline{\includegraphics[width=\columnwidth]{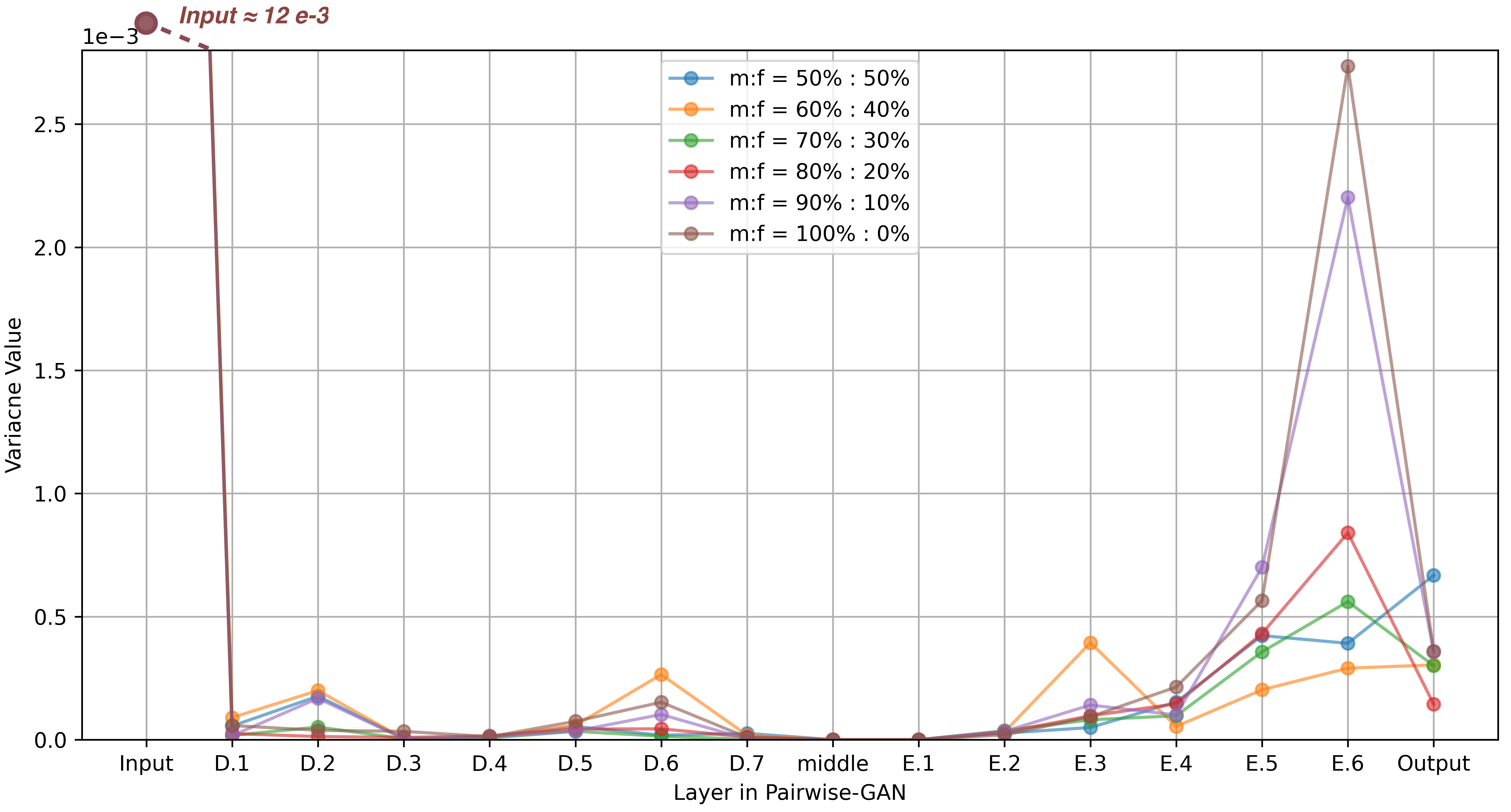}}
\caption{Variance Changes in Each Layer of Pairwise-GAN}
\label{variance}
\end{figure}

In the third stage, we evaluate which layers are vulnerable to biased data by calculating the variance of each layer. The results are shown in Fig.~\ref{variance}. The x-direction represents different layers of Pairwise-GAN from right to left, consisting of one input layer, seven decoder layers, one middle layer, and seven encoder layers. The variance of the last layer is also the variance of the output layer. The y-direction indicates variance values. The six lines in plot reflect the variance changes of the six pre-trained models.

The variance value decreases from $0.012$ to $0$ from the input layer to the middle layer. The value increased to around $0.0005$ in the encoder. More importantly, there are five points in the diagram where six pre-trained models have a large substantial difference: decoder layer $2$, decoder layer $6$, encoder layer $3$, encoder layer $5$, and encoder layer $6$. In particular, the fifth encoder layer and sixth encoder layer directly lead to the biases to the output. More specifically, biased models behave with very high variance in the sixth encoder. This indicates that as deconvolution operations in this layer map the vector from the last layer into a broader space, this contains more noise. Furthermore, we can also conclude that only some layers in Pairwise-GAN are significantly vulnerable to biased data.

Based on the clear results above, we conducted the fourth experiment, to attempt to make the model more stable to the biased data by modifying the generator's architecture in Pairwise-GAN. In the original generator, there exist connections between the shallow layers of the decoder and deep layers of the encoder. In detail, both regular features and biased noise can pass through the skip-connection, which cause an agglomeration effect. For instance, biases in the sixth encoder can be tracked to come from the encoder layer $5$ and decoder layer $5$. Furthermore, the variance differences between middle layers is lower compared to other layers. It is better for the encoder to receive inputs from low-feature layers rather than encoder layers. 

In the next step, evaluations of the modified model were analyzed by following the template in the first experiment. The results are shown in Table~\ref{mm_performance}. Compared to the previous model, the modified model can predict a face image from a female side-pose image even trained on a dataset with all-male images. However, there is only a relatively small improvement in the gender match rate. If the pre-trained model receives a female side-pose image, there is a $97.7\%$ probability that the predicted frontal image is a male image. In summary, the modified model has reduced vulnerability in the recovery rate and gender match rate. The model trained on an extremely biased dataset still can generate a human face; however, we had not yet found a method for models to resist all bias.

\begin{table}
\centering
\caption{Face recovery rate and gender match rate}
\label{mm_performance}
\begin{tabular}{c|c|cccccc}
\hline\hline
\multicolumn{2}{c|}{~\textbf{Male in train-set(\%)}}   & 50   & 60   & 70   & 80   & 90   & 100  \\
\multicolumn{2}{c|}{~\textbf{Female in train-set(\%)}} & 50   & 40   & 30   & 20   & 10   & 0    \\ \hline
\textit{Face}                  & male         & 0.99 & 1    & 1    & 1    & 1    & 1    \\ \cline{2-2}
\textit{Recovery Rate }      & female       & 1    & 1    & 1    & 1    & 1    & 1    \\ \hline
\textit{Gender}                & male         & 0.96 & 0.90 & 0.94 & 0.96 & 0.99 & 1    \\ \cline{2-2}
\textit{Match Rate }        & female       & 0.74 & 0.84 & 0.79 & 0.63 & 0.30 & 0.02 \\ \hline\hline
\end{tabular}
\end{table}

\subsection{Results on the biased filter study}
\label{sec:result4}

\begin{figure}[htbp]
\centerline{\includegraphics[width=0.95\columnwidth]{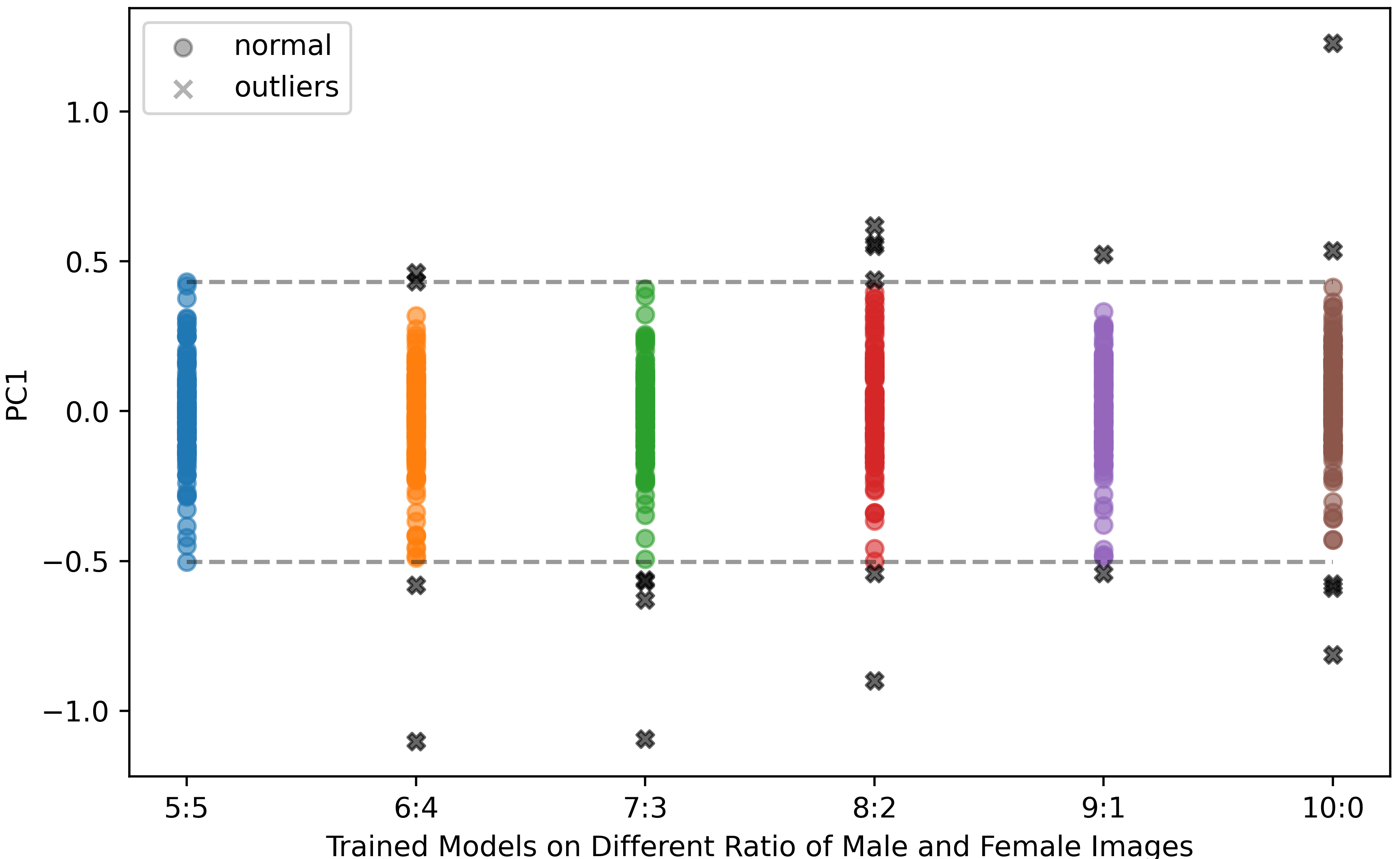}}
\caption{Scatter Filters in $Encoder - 6$. A Circle point refers to a filter that is normal; a crossing point refers to a filter is outliers.}
\label{filter6ana}
\end{figure}

We designed a filter analysis algorithm to investigate the filters on $6$ pre-trained Pairwise-GANs from experiment $B \ \& \ D$, where they were trained on the different ratios of male and female images. From the fig.\ref{filter6ana}, filters from model trained on unbiased dataset are symmetrically distributed on both sides of $0 \ PCA \ value$ with few outliers. Thus, we treated the range of filters in unbiased model as the reference standard to detect outliers. Compared to filters in the unbiased pre-trained model, filters from other pre-trained models tend to have narrow distribution and more outliers. Besides, filters from high-variance layers have more outliters, such as decoder layer $5$ and encoder layer $6$ (Fig.~\ref{filter6ana}). In the meantime, we also found that PCA distribution of filters from extremely biased models were close to filters trained on unbiased models. 

\begin{algorithm}[h] 
	\caption{Filter Analysis} 
	\label{FAalgorith} 
	\begin{algorithmic}[1]
		\STATE {\bfseries Input:} \\
		$M$: load six pre-trained Pairwise-GAN \\
        \REPEAT
            \STATE initialize placeholder $layer \ l_{j}$
            \STATE initialize placeholder $data$  
    		\FOR {model $m_{i}$ $\in$ $M$}
    		    \STATE Load $l_{j}$ in $m_{i}$, get filter $f_{j}$ from $l_{j}$
    		    \STATE where $f_{j}:[height, width, in\_cha, out\_cha]$
    		    \STATE Average pooling on $in\_cha$ dimension
    		    \STATE Flattens $f_{j}$ with shape $[height\times width, out\_cha]$
    		    \STATE Append $f_{j}$ in $data$
    		\ENDFOR
    	    \STATE get $data:$ $[height\times width, out\_cha_i \times 6]$
    	    \STATE perform PCA on $data$ get $[1, out\_cha_i \times 6]$
    	    \STATE Scatter $data$: x-dir: 6 model in $l_{j}$; y-dir: PCA value in top direction; point: top PCA feature of each filter
		\UNTIL {all layers in encoder and decoder have been iterated }
	\end{algorithmic} 
\end{algorithm}

\section{Conclusion}

Inspired by the study on biases in the Google translate system and PULSE (facial synthesis), we explore biases of facial synthesis. There is very limited research literature on generation models with respect to group biases. We conducted experiments on exploring gender bias in the face frontalization task. More specifically, we selected for investigation two representative models: Pix2Pix and Pairwise-GAN. At the beginning stage, we explored the biased behaviors for these models from the gender perspective. To further explore the model behaviors, more latent space in test-sets was evaluated. Based on these findings, we investigated the cause of model vulnerability to biased data by studying the latent space and kernel of the generator, and proposed suggestions that can strengthen the model's resistance to biased data. 

In the first two experiments, we conclude that both Pix2Pix and Pairwise-GAN are "fair" algorithms in gender bias, but they are vulnerable to biased training data. This is commensurate with GAN theory, as the distribution transaction functions are learnt during the training process. If the training-set contains biases, the transaction function will introduce the biases to the pre-trained models. Furthermore, we found that biased pre-trained models always predict biased results even if the test-set is a different distribution from the training-set. This finding is similar to human beings; biased viewpoints tend to let people fall into a biased habits.

In conclusion to the second stage of experiments, our purpose was to find the reason for the biases. Through calculating the variance in latent space, we concluded that some layers of Pairwise-GAN are particularly vulnerable to biased training data. We deduced that skip-connection in the generator can transmit the layer biases from the decoder into the encoder, therefore, we proposed a modified generator of Pairwise-GAN by cutting off three skip-connections. The model behaves in a more stable fashion where one biased behavior (failure in generation) is eliminated, and the other (gender balance) is improved.

In further, other facial datasets will involve in validating our findings. We also plan to extend the conclusions from Pix2Pix and Pairwise-GAN to other conditional-GANs, proposing a general conclusion. Furthermore, we still have confidence in minimizing known bias under a limited size dataset.





\begin{thebibliography}{00}
\bibitem{b1}N. Mehrabi, F. Morstatter, N. Saxena, K. Lerman and A., Galstyan, A survey on bias and fairness in machine learning, arXiv preprint arXiv:1908.09635, 2019.
\bibitem{b2}M.O. Prates, P.H. Avelar and L.C. Lamb, Assessing gender bias in machine translation: a case study with google translate, Neural Computing and Applications, pp.1-19, 2019.
\bibitem{b3}X. Shen, J. Plested, Y. Yao and T. Gedeon, Pairwise-GAN: Pose-Based View Synthesis Through Pair-Wise Training, In International Conference on Neural Information Processing, pp. 507-515, Springer, November 2020. 
\bibitem{b4}P. Isola, J.Y. Zhu, T. Zhou and A.A. Efros, Image-to-image translation with conditional adversarial networks, In Proceedings of the IEEE conference on computer vision and pattern recognition, pp. 1125-1134, 2017.
\bibitem{b5}J.Y. Zhu, T. Park, P. Isola and A.A. Efros, Unpaired image-to-image translation using cycle-consistent adversarial networks, In Proceedings of the IEEE international conference on computer vision, pp. 2223-2232, 2017.
\bibitem{b6}Y. Wei, M. Liu, H. Wang, R. Zhu, G. Hu and W. Zuo, Learning Flow-based Feature Warping for Face Frontalization with Illumination Inconsistent Supervision, In European Conference on Computer Vision, pp. 558-574, Springer, August 2020.
\bibitem{b7}H. Suresh, J.V. and Guttag, A framework for understanding unintended consequences of machine learning, arXiv preprint arXiv:1901.10002, 2019.
\bibitem{b8}A. Olteanu, C. Castillo, F. Diaz and E. Kiciman, Social data: Biases, methodological pitfalls, and ethical boundaries, Frontiers in Big Data, vol.2, p.13, 2019.
\bibitem{b9}S. Barocas, M. Hardt and A. Narayanan, Fairness in machine learning, NIPS Tutorial, vol 1, 2017.
\bibitem{b10}R. Huang, S. Zhang, T. Li and R. He, Beyond face rotation: Global and local perception gan for photorealistic and identity preserving frontal view synthesis, In Proceedings of the IEEE International Conference on Computer Vision, pp. 2439-2448, 2017.
\bibitem{b11}W. Zhuang, L. Chen, C. Hong, Y. Liang and K. Wu, FT-GAN: Face Transformation with Key Points Alignment for Pose-Invariant Face Recognition, Electronics, vol 8, p.807. 2019.
\bibitem{b12}K. Crawford, The trouble with bias, In Conference on Neural Information Processing Systems, December 2017. 
\bibitem{b13}J. Buolamwini and T. Gebru, Gender shades: Intersectional accuracy disparities in commercial gender classification, In Conference on fairness, accountability and transparency, pp. 77-91, January 2018.
\bibitem{b14}S. Nagpal, M. Singh, R. Singh and M. Vatsa, Deep learning for face recognition: Pride or prejudiced?, arXiv preprint arXiv:1904.01219, 2019
\bibitem{b15}J.G. Cavazos, P.J. Phillips, C.D. Castillo and A.J. O’Toole, Accuracy comparison across face recognition algorithms: Where are we on measuring race bias?, IEEE Transactions on Biometrics, Behavior, and Identity Science, 2020.
\bibitem{b16}S.Yucer, S. Akçay, N. Al-Moubayed and T.P. Breckon, Exploring Racial Bias within Face Recognition via per-subject Adversarially-Enabled Data Augmentation, In Proceedings of the IEEE/CVF Conference on Computer Vision and Pattern Recognition Workshops, pp. 18-19, 2020.
\bibitem{b17}S. Shankar, Y. Halpern, E. Breck, J. Atwood, J. Wilson and D. Sculley, No classification without representation: Assessing geodiversity issues in open data sets for the developing world, arXiv preprint arXiv:1711.08536, 2017.
\bibitem{b18}Y. Zhang, M. Shao, E.K. Wong and Y. Fu, Random faces guided sparse many-to-one encoder for pose-invariant face recognition. In Proceedings of the IEEE International Conference on Computer Vision, pp. 2416-2423, 2013.
\bibitem{b19}M. Kan, S. Shan, H. Chang and X. Chen, Stacked progressive auto-encoders (spae) for face recognition across poses, In Proceedings of the IEEE conference on computer vision and pattern recognition, pp. 1883-1890, 2014.
\bibitem{b20}T. Hassner, S. Harel, E. Paz and R. Enbar, Effective face frontalization in unconstrained images, In Proceedings of the IEEE conference on computer vision and pattern recognition, pp. 4295-4304, 2015.
\bibitem{b21}X. Yin, X, Yu, K. Sohn, X. Liu and M. Chandraker, Towards large-pose face frontalization in the wild, In Proceedings of the IEEE international conference on computer vision, pp. 3990-3999, 2017.
\bibitem{b22}Y. Tian, X. Peng, L. Zhao, S. Zhang and D.N. Metaxas, CR-GAN: learning complete representations for multi-view generation, arXiv preprint arXiv:1806.11191, 2018.
\bibitem{b23}M. Pieters and M. Wiering, Comparing generative adversarial network techniques for image creation and modification, arXiv preprint arXiv:1803.09093, 2018.
\bibitem{b24}T. White, Sampling generative networks, arXiv preprint arXiv:1609.04468, 2016.
\bibitem{b25}D. Bau, J.Y. Zhu, H. Strobelt, B. Zhou, J.B. Tenenbaum, et al, Gan dissection: Visualizing and understanding generative adversarial networks, arXiv preprint arXiv:1811.10597, 2018.
\bibitem{b26}H.Y. Chang, Z. Wang and Y.Y. Chuang, Domain-Specific Mappings for Generative Adversarial Style Transfer, arXiv preprint arXiv:2008.02198, 2020.
\bibitem{b27}A. Liu, S. Ginosar, T. Zhou, A.A. Efros and N. Snavely, Learning to Factorize and Relight a City, arXiv e-prints, pp.arXiv-2008, 2020.
\bibitem{b28}Y. Yin, S. Jiang, J.P. Robinson and Y. Fu, Dual-attention GAN for large-pose face frontalization, arXiv preprint arXiv:2002.07227, 2020.

\end{thebibliography}
\end{document}